\documentclass[conference]{IEEEtran}
\IEEEoverridecommandlockouts

\usepackage{pifont}

\usepackage{hyperref}
\hypersetup{hypertex=true,
            colorlinks=true,
            linkcolor=blue,
            anchorcolor=blue,
            citecolor=blue}
            
\usepackage{cite}
\usepackage{amsmath,amssymb,amsfonts}
\usepackage{algorithmic}
\usepackage{graphicx}
\usepackage{pifont}
\usepackage{textcomp}
\usepackage{xcolor}
\def\BibTeX{{\rm B\kern-.05em{\sc i\kern-.025em b}\kern-.08em
    T\kern-.1667em\lower.7ex\hbox{E}\kern-.125emX}}

\author{
\IEEEauthorblockN{Xiangfei Qiu$^{1*}$, 
Xiuwen Li$^{1*}$,
Ruiyang Pang$^{1*}$,
Zhicheng Pan$^{1*}$,
Xingjian Wu$^{1*}$,
Liu Yang$^{1*}$,
Jilin Hu$^{1}$,\\
Yang Shu$^{1}$,
Xuesong Lu$^{1}$,
Chengcheng Yang$^{1}$,
Chenjuan Guo$^{1}$,
Aoying Zhou$^{1}$,
Christian S. Jensen$^{2}$ and
Bin Yang$^{1}$
} 
\IEEEauthorblockA{$^1$\textit{School of Data Science and Engineering, East China Normal University, Shanghai, China} \\ $^2$\textit{Department of Computer Science, Aalborg University, Aalborg, Denmark} \\
}
}

\begin{document}

\title{EasyTime: Time Series Forecasting Made Easy}

\maketitle

\begin{abstract}
Time series forecasting has important applications across diverse domains. EasyTime, the system we demonstrate, facilitates easy use of time-series forecasting methods by researchers and practitioners alike. First, EasyTime enables one-click evaluation, enabling researchers to evaluate new forecasting methods using the suite of diverse time series datasets collected in the preexisting time series forecasting benchmark (TFB). This is achieved by leveraging TFB’s flexible and consistent evaluation pipeline. Second, when practitioners must perform forecasting on a new dataset, a nontrivial first step is often to find an appropriate forecasting method.  EasyTime provides an Automated Ensemble module that combines the promising forecasting methods to yield superior forecasting accuracy compared to individual methods. Third, EasyTime offers a natural
language Q\&A module leveraging large language models. Given a question like “Which method is best for long term forecasting on time series with strong seasonality?”, EasyTime converts the question into SQL queries on the database of results obtained by TFB and then returns an answer in natural language and charts. By demonstrating EasyTime, we intend to show how it is possible to simplify the use of time-series forecasting and to offer better support for the development of new generations of time series forecasting methods.
\end{abstract}

\begin{IEEEkeywords}
time series forecasting, platform.
\end{IEEEkeywords}

\section{Introduction}
Time series, time-ordered sequences of data points, are generated in a variety of domains, including finance, transportation, and health~\cite{wu2024catch,wu2024autocts++,gao2024diffimp,hu2024multirc}. Time series forecasting~(TSF) is important functionality that enables leveraging time series for making predictions and making decisions~\cite{wu2023timesnet,qiu2025duet}. In finance, TSF is employed to predict stock prices, exchange rates, and interest rates, aiding investors in formulating investment strategies and managing risks. In meteorology, TSF is utilized for weather forecasting, assisting people in planning activities that are influenced by the weather. Not surprisingly, TSF is an active field of research field that has produced numerous diverse methods ranging from statistical methods to deep learning methods. However, it remains difficult for researchers as well as practitioners with different skill levels to evaluate new methods, identify appropriate methods when having to perform forecasting on new datasets, and get answers to relevant questions related to TSF.

\textbf{Challenge 1. It is difficult to comprehensively evaluate a new TSF method. } 
To comprehensively evaluate a method, several key aspects should be considered. First, the evaluation should encompass datasets from diverse domains, ensuring coverage of different dataset characteristics. Second, a wide range of existing methods should be included for comparison to capture performance differences among methods. Third, different evaluation strategies, such as fixed-window and rolling forecasting, should be employed when assessing method performance. 
Fourth, it is desirable to use multiple evaluation metrics to get a nuanced understanding of method performance. Fifth, attention should be paid to the consistency of the evaluation setup, e.g., the partition in training/validation/testing data, the choice of normalization techniques, the hyperparameter settings, and the use of the ``drop last'' operation~\cite{qiu2024tfb,li2024foundts}. However, it is time-consuming to cover all these aspects, and few studies consider all aspects. The result can be partial evaluations and misleading conclusions.

\begin{figure*}[t!]
    \centering
    \includegraphics[width=1\linewidth]{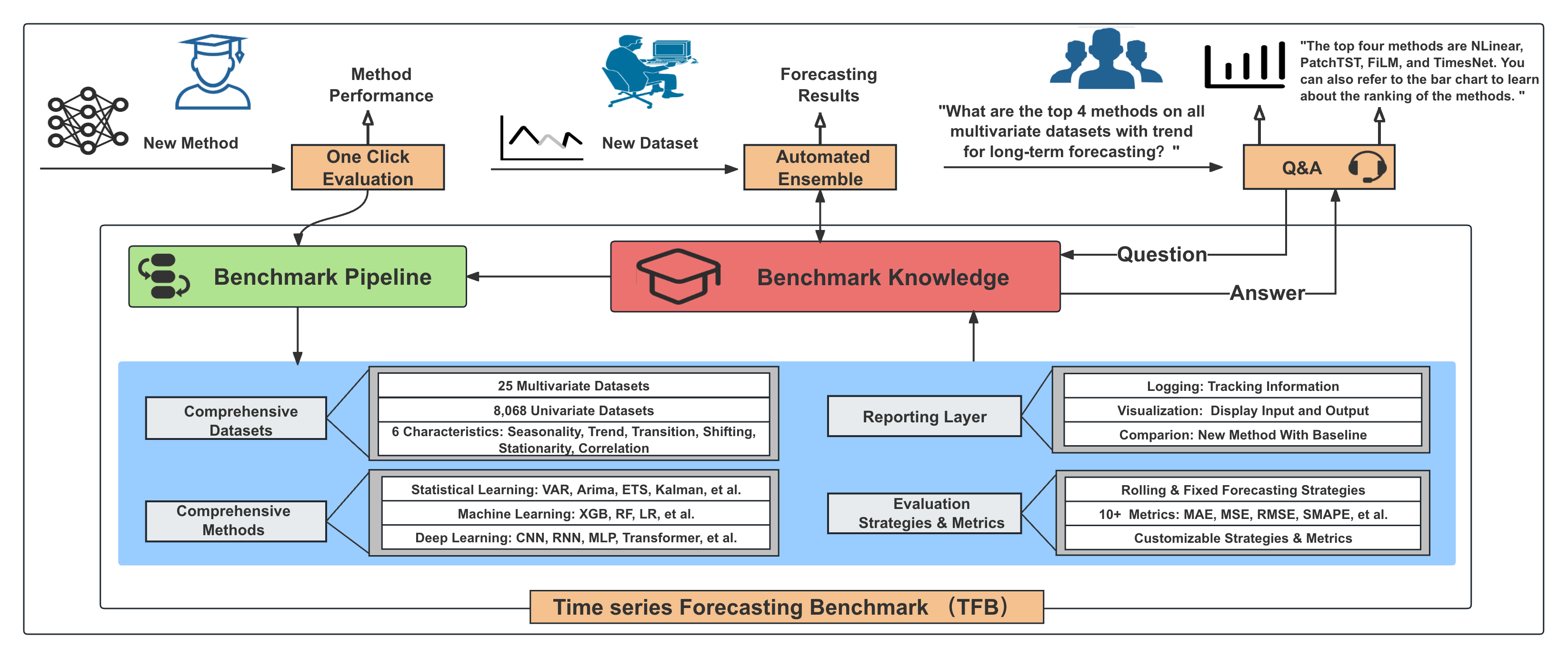}
    \vspace{-5mm}
    \caption{EasyTime Overview.}
    \label{fig: Architecture}
    \vspace{-3mm}
\end{figure*}


\textbf{Challenge 2. It is difficult for practitioners to determine which forecasting methods to use on new datasets.} This challenge arises due to several factors. First, different methods exhibit varying capabilities on datasets depending on their distinct characteristics---there is no single best solution. Consequently, given a new dataset, selecting the most suitable method is important but also challenging.  Second, new datasets may exhibit unique characteristics and patterns compared to previous datasets, potentially leading to no existing method being satisfactory. Analyzing new datasets requires a deep understanding of the data, demanding advanced domain knowledge and data analysis skills. 

\textbf{Challenge 3. It is difficult for practitioners to get answers to relevant questions related to TSF.}
With the development of large language models~(LLMs), it is realistic to envision AI agents that can interact with practitioners to answer their questions about TSF. For beginners, answers to questions like "Is the Transformer or LSTMs better for time series with trends?" may facilitate their learning. While business people can get answers to questions formulated in natural language like ``Will the sales in Shanghai increase next month?'' TSF methods generally cannot be queried using natural language interfaces, and general LLMs may not have the specific knowledge of TSF methods.

To address these challenges, we present EasyTime. The system's functionality is implemented in four modules: the \textit{Time series Forecasting Benchmark}, the \textit{One Click Evaluation Module}, the \textit{Automated Ensemble Module}, and the \textit{Q\&A Module}, as shown in Figure~\ref{fig: Architecture}. EasyTime leverages TFB\footnote{\url{https://github.com/decisionintelligence/TFB}}, a comprehensive and fair Time series Forecasting
Benchmark~\cite{qiu2024tfb}, which incorporates a suite of datasets spanning dozens of application domains to represent dataset characteristics comprehensively; and it encompasses a diverse range of state-of-the-art methods that include statistical learning, machine learning, and deep learning methods. It also provides support for different evaluation strategies and metrics thus enabling comprehensive evaluation of forecasting methods. In addition, TFB includes a consistent and flexible pipeline for evaluating method performance. This allows researchers to evaluate the performance of new methods with just one-click. 
Moreover, TFB has accumulated a large number of benchmarking results from evaluating 30+ methods on 8,000+ time series. 
%
%
These results are highly valuable as they offer comprehensive insight into the performance of forecasting methods in diverse settings. Utilizing these results as a knowledge base, when facing a new time series data set, it is possible to select the most relevant methods and combines them into an ensemble model to enable accurate forecasting. 
Furthermore, we augment existing large language models with the benchmark knowledge 
to facilitate natural langauge Q\&A on time series forecasting. 

The rest of the paper is structured as follows. We introduce EasyTime’s architecture in Section~\ref{EasyTime ARCHITECTURE}, and provide the demonstration outline of EasyTime 
in Section~\ref{SYSTEM DEMONSTRATION}.

\section{EasyTime ARCHITECTURE}
\label{EasyTime ARCHITECTURE}

\subsection{Time series Forecasting Benchmark}
TFB encompasses a data layer, a method layer, an evaluation layer, a reporting layer, a benchmark pipeline, and benchmark knowledge---see Figure~\ref{fig: Architecture}. The data layer includes 25 multivariate datasets and 8,068 univariate datasets from 10 different domains: traffic, electricity, energy, the environment, nature, economic, stock markets, banking, health, and the web. Moreover, the datasets are chosen to provide good representation of characteristics such as Seasonality, Trend, Transition, Shifting, Stationarity, and Correlation. The method layer offers a flexible interface that facilitates the inclusion of statistical learning, machine learning, deep learning, and foundation time series forecasting methods. It also ensures compatibility with other third-party TSF libraries, such as Darts~\cite{herzen2022darts} and TSLib~\cite{wu2023timesnet}. Users can easily integrate their own forecasting methods implemented using the third-party libraries into the benchmark. The evaluation layer supports a wide range of evaluation strategies and metrics, including both fixed-window and rolling forecasting strategies. It includes well-recognized evaluation metrics and allows for the use of customized metrics to thoroughly evaluate performance. The reporting layer includes a logging system for tracking experimental information and  supports visualization of time series inputs and forecasting results. 
%
The benchmark pipeline facilitates standardized dataset processing and splitting, modeling training and testing, as well as unified post-processing. 
When users 
include their methods into 
the method layer along with a configuration file, they can automatically run the pipeline to obtain performance results. The benchmark knowledge consists of the meta-information of both datasets and methods, and also the benchmarking experiment results of 30+ TSF methods on 8,000+ time series.




\subsection{One Click Evaluation}
When researchers wish to evaluate new or old methods in new or old forecasting scenarios, EasyTime's one-click evaluation comes in handy. EasyTime is scalable and user-friendly. 
Researchers only need to integrate their method into the TFB pipeline; then the performance of their method under different experimental settings and evaluation strategies can be evaluated with just one click. EasyTime also offers to run a method on all existing datasets with one click. Next, when researchers encounter new forecasting scenarios, such as the need to predict new forecasting horizons or perform rolling forecasts on univariate datasets, they just need to modify the system's configuration file information. This includes changing the evaluation strategy to rolling forecasting, adjusting the forecasting horizons, etc. Then, they can utilize the one-click evaluation.


\begin{figure}[t]
    \centering
    \includegraphics[width=1\linewidth]{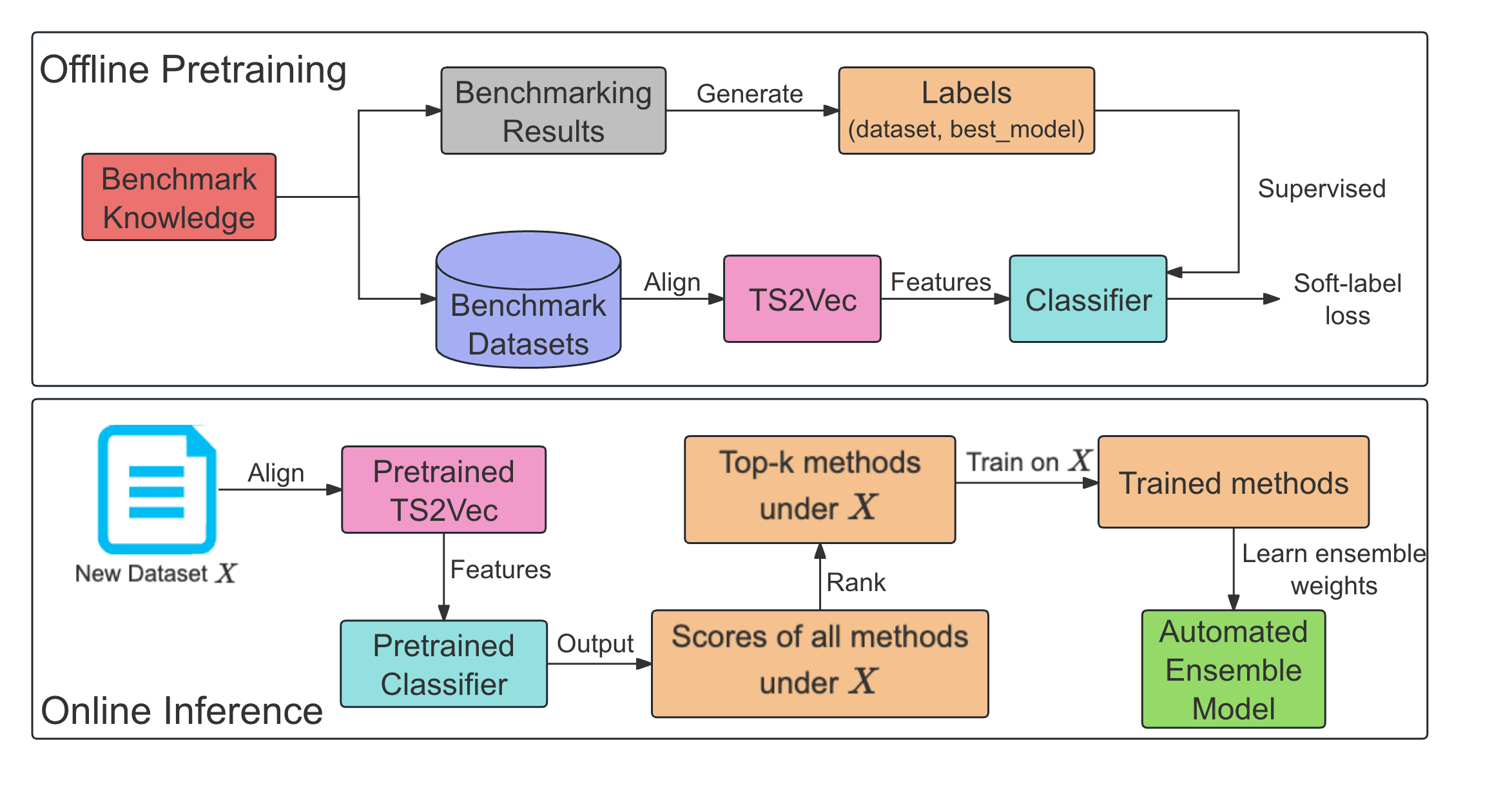}
    \vspace{-8mm}
    \caption{Automated Ensemble Overview.}
    \label{fig: AutoML}
    \vspace{-5mm}
\end{figure}

\subsection{Automated Ensemble}

EasyTime employs the Automated Ensemble Module to help practitioners, e.g., data scientists, build an appropriate method for a specific TSF dataset. As one of the main focus of the demonstration, the Automated Ensemble Module enables: \textit{offline pretraining} and \textit{online inference}---see Figure~\ref{fig: AutoML}. During \textit{offline pretraining}, we first train an unsupervised representation learning model TS2Vec~\cite{yue2022ts2vec} to extract features of time series. Then, to learn the correlations between features and model performance, we train a classifier on the basis of evaluations of 30+ high-performance UTSF methods on 8,000+ time series. By using the \textit{soft-label} loss~\cite{yao2023simplets}, the classifier outputs a probability ranking of methods. Leveraging these pre-trained components, EasyTime automates the modeling process. More specifically, when a new dataset $X$ arrives in the online inference phase, the pre-trained TS2Vec first extracts features from $X$. The pre-trained classifier then outputs top-$k$ methods based on the features. Next, EasyTime trains the top-$k$ candidate methods on the training part of dataset $X$ and learns the ensemble weights on the validation part of $X$ such that it fits the best to $X$. 
Finally, the ensemble model performs accurate forecasting on $X$.



\subsection{Natural Language Q\&A}

\begin{figure}[t]
    \centering
    \includegraphics[width=1.0\linewidth]{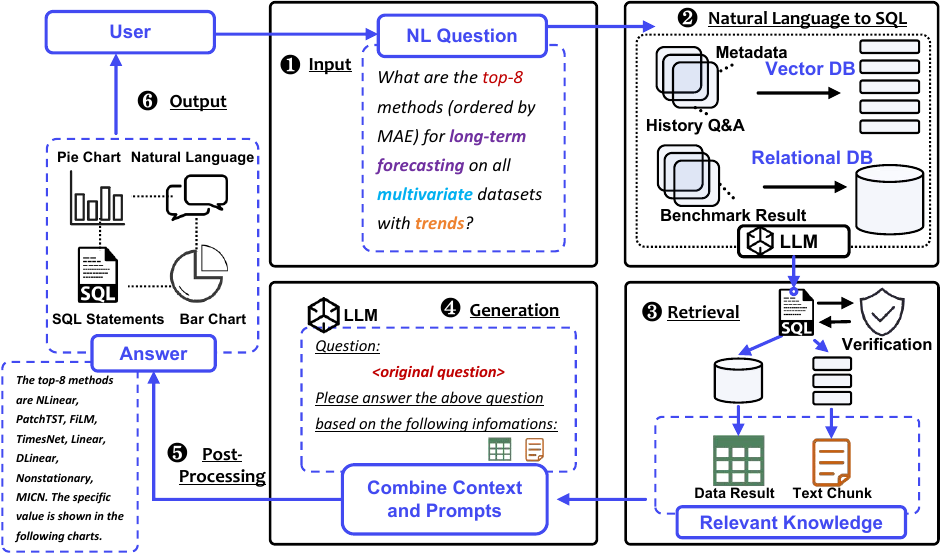}
    \vspace{-6mm}
    \caption{Workflow of Natural Language Q\&A.}
    \vspace{-5mm}
    \label{xxx-qa}
\end{figure}

EasyTime offers a natural language Q\&A
module that combines LLMs with the benchmark knowledge. As shown in Figure~\ref{xxx-qa}, the workflow of the Q\&A module is as follows: \ding{182}~\textbf{Input:} Users ask natural language (NL) questions about time series forecasting. 
\ding{183}~\textbf{NL2SQL:} EasyTime first combines pre-stored benchmark metadata, Q\&A history, with the current user's natural language query. Then, it employs the LLM to effectively generate the corresponding SQL statements. \ding{184}~\textbf{Retrieval:} The SQL statements 
are first verified for correctness before they are executed on the comprehensive knowledge base. This two-step approach ensures the accuracy and reliability of the query execution, thereby maintaining the integrity of the data while delivering the most appropriate output. \ding{185}~\textbf{Generation:} EasyTime populates the initial user question and retrieved auxiliary knowledge results into a predefined prompt template and employs the LLM for inferencing. \ding{186}~\textbf{Post-Processing:} The result of the LLM inferencing is presented in user-friendly natural language. Additionally, the data query results related to the user's question are provided as structured data outputs compatible with various types of charts. \ding{187}~\textbf{Output:} EasyTime returns all obtained results, including natural language responses, charts, SQL statements, and data query results, to the frontend for rendering, thereby providing users with interactive and visualized responses. 

\section{SYSTEM DEMONSTRATION}
\label{SYSTEM DEMONSTRATION}

\begin{figure*}[t]
    \centering
    \includegraphics[height=7.8cm,width=18cm]{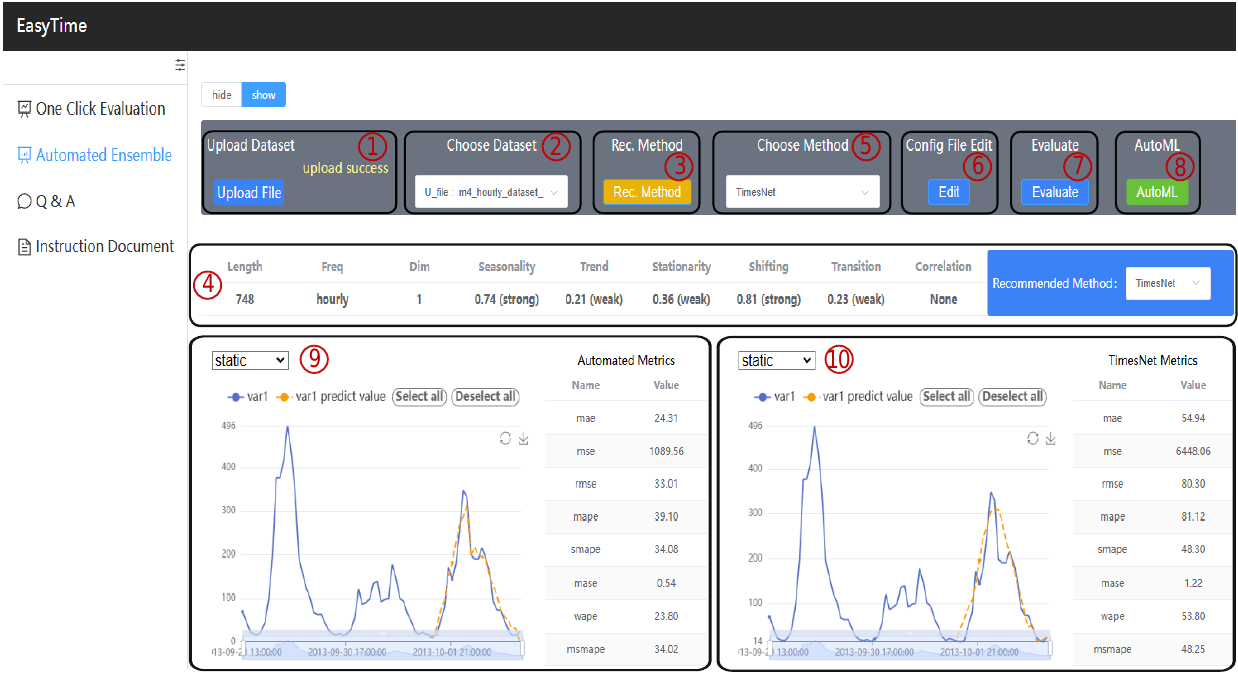}
    \vspace{-1mm}
    \caption{An Example of Method Recommendation, Automated Ensembles, and Forecasts Visualizations. }
    \label{fig:pattern_eg}
    \vspace{-3mm}
\end{figure*}

\begin{figure}[t]
    \centering
    \includegraphics[width=1\linewidth]{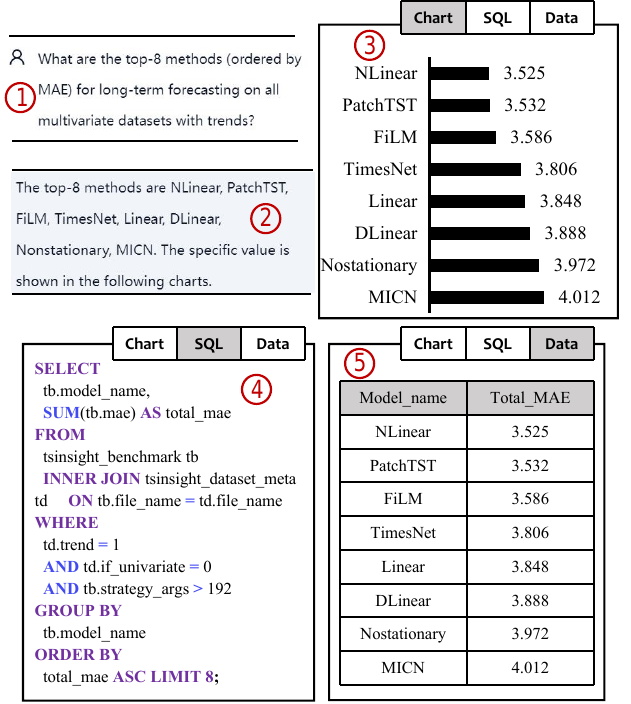}
    \vspace{-3mm}
    \caption{An Example of EasyTime Q\&A.}
    \label{Screenshot}
    \vspace{-5mm}
\end{figure}



\noindent
\textbf{S1. One Click Evaluation.} We initially demonstrate the programmatic configuration of EasyTime for one click evaluation using the web frontend. First, users only need to follow the specifications of our pipeline method embedding to easily embed their methods into the pipeline. Second, 
EasyTime simplifies the configuration of complex evaluation tasks, making these easy to accomplish. Users need only edit the configuration file in the web frontend, thus achieving one click evaluation.
We show the variety of existing methods and encourage researchers to perform their own systematic evaluations.

\vspace{3pt}\noindent
\textbf{S2. Automated Ensemble.} 
We demonstrate that it is easy to form an ensemble method to do forecasting on new time series with our interactive web frontend---see Figure~\ref{fig:pattern_eg}. 
At first, the user can either upload their data~(with the Upload Dataset button~(see label~\ding{182})) or choose a time series from the benchmark~(with the Choose Dataset button~(see label~\ding{183})). 
Then, the user only needs to click the ``Recommend Method'' button~(see label~\ding{184}), and the characteristics of the time series and the recommended method are displayed~(see label~\ding{185}). 
Next, the user can choose a method~(recommended or any other individual method supported by EasyTime) to evaluate the time series~(see label~\ding{186}). Meanwhile, the user can also edit the configuration file~(such as selecting the forecasting horizon and evaluation strategy) to configure the forecasting scenario~(see label~\ding{187}). 
After that, the user only needs to click the ``Evaluate'' button~(see label~\ding{188}) to start the evaluation process. 
Meanwhile, the user can click the ``AutoML'' button~(see label~\ding{189}) to ensemble a forecasting method that can fit the selected time series the best automatically. 
Finally, the visualizations and evaluation metrics of forecasts on both ensemble and chosen methods are displayed~(see labels~\ding{190} and~\ding{191}), respectively.


\vspace{3pt}\noindent
\textbf{S3. Natural Language Q\&A.} We demonstrate the natural language Q\&A capability of EasyTime---see Figure~\ref{Screenshot}. Users can obtain historical benchmark results directly by \ding{182}~asking questions, e.g., ``\textit{What are the top-8 methods (ordered by MAE) for long-term forecasting on all multivariate datasets with trends?}'' The results returned encompass \ding{183}~natural language responses, \ding{184}~Visualized charts~(bar charts, line charts, pie charts, etc.), \ding{185}~the corresponding SQL statements (to ensure the correctness of the underlying logic), and \ding{186}~benchmark result data table. This Q\&A feature further improves ease-of-use of EasyTime, allowing users to identify the strengths of different TFS methods most naturally and to consider decision-making in real-world scenarios.

We hope that our demonstration can shed lights on how we can simplify the use of of time series forecasting and better support the development of the next generation of time series forecasting.

\bibliographystyle{IEEEtranS}
\IEEEtriggeratref{65}
\bibliography{sample}

\end{document}